\documentclass[journal,transmag]{IEEEtran}

\usepackage[utf8]{inputenc} 
\usepackage[T1]{fontenc}    
\usepackage{hyperref}       
\usepackage{url}            
\usepackage{booktabs}       
\usepackage{amsfonts}       
\usepackage{nicefrac}       
\usepackage{microtype}      

\usepackage{amsmath}
\usepackage{graphicx}
\usepackage{setspace}

\title{Exploiting Spatio-Temporal Structure with Recurrent Winner-Take-All Networks}

\author{
  Eder Santana \\
  Department of Electrical and Computer Engineering\\
  University of Florida\\
  Gainesville, FL 32611 \\
  \texttt{edersantana@ufl.edu} \\
  \And
  Matthew Emigh \\
  University of Florida \\
  Gainesville, FL 32611 \\
  matt@cnel.ufl.edu \\
  \AND
  Pablo Zegers \\
  Universidad de los Andes \\
  Facultad de Ingenieria y Ciencias Aplicadas \\
  pablozegers@gmail.com \\
  \And
  Jose Principe \\
  University of Florida \\
  Gainesville, FL 32611 \\
  principe@cnel.ufl.edu \\
}

\begin{document}
\author{\IEEEauthorblockN{Eder Santana\IEEEauthorrefmark{1},
Matthew Emigh\IEEEauthorrefmark{1},
Pablo Zegers\IEEEauthorrefmark{2}, 
Jose C Principe\IEEEauthorrefmark{1},~\IEEEmembership{Fellow,~IEEE}}
\IEEEauthorblockA{\IEEEauthorrefmark{1}Department of Electrical and Computer Engineering,
University of Florida, Gainesville, FL 32611 USA}
\IEEEauthorblockA{\IEEEauthorrefmark{2} Facultad de Ingenieria y Ciencias Aplicadas, Universidad de los Andes, Chile}
\thanks{This paper was partially funded by University of Florida Graduate Scholarship and ONR N00014-14-1-0542. Corresponding author: E. Santana (DM: https://twitter.com/edersantana).}}

\markboth{Under Review}%
{Shell \MakeLowercase{\textit{et al.}}: Bare Demo of IEEEtran.cls for IEEE Transactions on Magnetics Journals}
\IEEEtitleabstractindextext{%
\begin{abstract}
  We propose a convolutional recurrent neural network, with Winner-Take-All dropout for high dimensional unsupervised feature learning in multi-dimensional time series. We apply the proposedmethod for object recognition with temporal context in videos and obtain better results than comparable methods in the literature, including the Deep Predictive Coding Networks previously proposed by Chalasani and Principe.Our contributions can be summarized as a scalable reinterpretation of the Deep Predictive Coding Networks trained end-to-end with backpropagation through time, an extension of the previously proposed Winner-Take-All Autoencoders to sequences in time, and a new technique for initializing and regularizing convolutional-recurrent neural networks.

\end{abstract}

\begin{IEEEkeywords}
deep learning, unsupervised learning, convolutional recurrent neural networks, winner-take-all, object recognition
\end{IEEEkeywords}}

\maketitle

\section{Introduction}
\label{sec:intro}
An elusive problem for both the cognitive and machine learning communities is the
precise algorithm by which the human sensory system interprets the continuous stream
of sensory inputs as stable perceptions of recognized objects and actions. In engineering,
object and action recognition are tackled with supervised learning; on the other hand
we have no evidence that the brain uses hard-wired or genetically evolved "supervisors"
for training recognition networks. We argue that even if our nervous system applies supervised learning,
the object classes should emerge in a self organizing way from experience and can be used as supervising labels.

It has been suggested that temporal information could be a exploited as a possible source of supervision\cite{Baker2014}. In this respect,
suppose an observer is moving around and looking at an object in the middle of a room. Given that the input visual
stream is continuous and that the object does not move, all the observed images of the object must
belong to related perceptions and be represented in similar ways. This smooth, temporally coherent representation
is biologically plausible and has been shown to self-organize V1-like Gabor filters when applied to learn image
transition representations in video \cite{hurri2002temporal}.

A machine learning application of this paradigm is to use temporal coherence
as a proxy for learning sensory representations without strong supervision or explicit labels \cite{Goroshin_2015_ICCV}\cite{Wang15}\cite{AgrawalCM15}.
One approach is a Bayesian formulation in which we assume the learning system builds an
internal model of the world $\tilde{p}(x_t; \theta)$ for explaining input streams $x_t$ using a system
parameterized by $\theta$. Predictive Coding proposes to adapt this model to reduce discrepancies between
predictions $\tilde{p}(x_t|x_{t-1}; \theta)$ and observations $p(x_t)$. Chalasani and Principe \cite{chalasani2013}
developed a hierarchical, distributed, generative architecture called Deep Predictive Coding Networks (DPCN)
that learns with free energy to build 
spatio-temporal shift-invariant representations of input video streams.
They showed that DPCNs learned features which can be used for classification, with competitive results albeit of being unsupervised.
One of the difficulties of this approach is the required inference even in the testing state, which makes it rather slow.

This paper is inspired in DPCNs but substitutes the top down inference step in DPCN by a recurrent 
convolutional encoder-decoder that predicts the next frame of the video, can be trained with backpropagation,
and does simples recall in test phase. The paper consists of three main contributions.
First, we evolved the self-organizing object recognition in video work of
Chalasani and Principe \cite{chalasani2013}\cite{chalasani2015context} by developing a scalable
counterpart to the DPCN architecture and algorithms. Second, we build our contributions on top of recent findings in convolutional winner-take-all
autoencoders \cite{makhzani2015winner} and convolutional-recurrent neural networks 
\cite{Liang_2015_CVPR}\cite{shi2015convrnn}. Thus, we extend the results of winner-take-all autoencoders to the time domain. We show results in video predictions, object recognition
and action recognition. Third, Luong et. al. \cite{Luong2015multitask} showed that RNNs benefit from unsupervised pre-training and multi-task learning. We show that our method can used as a pre-training technique for initializing convolutional RNNs. 

To present the guiding principles for the proposed architecture, in the next section we overview DPCNs and point its desirable features we would like to preserve and the undesirable features will like to substitute.

\section{Deep Predictive Coding Networks}
\label{sec:dpcn}

Assume a multi-dimensional time series (e.g., a video) $\mathbf{x}_t$. Chalasani and Principe \cite{chalasani2013}\cite{chalasani2015context} proposed a generative, dynamical, hierarchical model with sparse coefficients $\mathbf{s}_t$:
\begin{align}
\mathbf{s}_t &= \mathbf{A} \mathbf{s}_{t-1} + \mathbf{v}_t \\
\mathbf{x}_t &= \mathbf{C} \mathbf{s}_t + \mathbf{w}_t,
\end{align}
The sparsity of $\mathbf{s}_t$ is controlled by a nonlinear higher order statistical component $\mathbf{B} \mathbf{u}_t$, where $\mathbf{B}$ are the component weights and $\mathbf{u}_t$ is itself $L_1$-constrained to be sparse. This model can be stacked by generatively explaining, at layer $l$, the $\mathbf{u}_t$ from the layer below it: $\mathbf{u}_t^l = \mathbf{C} \mathbf{s}_t^{l+1} + \mathbf{w}_t^{l+1}$. In practice, this is accomplished by greedy layer-wise training.
DPCNs are trained with Expectation-Maximization (EM) using the following energy function:

\begin{equation}
    \begin{split}
        \mathcal{E}_t = \|\mathbf{x}_t - \mathbf{C} \mathbf{s}_t \|_2  + \lambda \|\mathbf{s}_t - \mathbf{A} \mathbf{s}_{t-1}\|_1  + \beta\|\mathbf{u}_t\|_1 \\
        + \sum_k^K | \mathbf{z}_{t}(k) \cdot \mathbf{s}_{t} (k) |,
    \end{split}
\end{equation}

\begin{equation}
    \mathbf{z}_t = \gamma_0 \frac{1+\exp(-\mathbf{B} \mathbf{u}_t)}{2},
\end{equation}
where the exponential function is applied to each element of the vector $-\mathbf{B} \mathbf{u}_t$, and $k$ represents the vector indices.

Thus, given parameters $\mathbf{C}$, $\mathbf{A}$ and $\mathbf{B}$, the DPCN algorithm searches for the $\mathbf{s}_t$ that best fits
an input $\mathbf{x}_t$. The term $\|\mathbf{s}_t - \mathbf{A} \mathbf{s}_{t-1}\|$ is a constraint that forces the solution to be as close as possible
to a linear update of the solution for the previous input frame $\mathbf{x}_{t-1}$. This is an $L_1$-slowness constraint
to force temporal smoothness in the representation. The constraint $\sum_k^K | \mathbf{z}_{t}(k) \cdot \mathbf{x}_{t} (k) |$ forces the solution
to be sparse, with sparsity level controlled by the higher level component $\mathbf{u}_t$. By doing so it also forces the network to
learn time and space invariant features in the variable $z_t(k)$. 

DPCNs can be extended to handle large images by substituting the projection matrices ${\mathbf{A}, \mathbf{B}}$ with convolutions and
the latent codes ${\mathbf{s}_t, \mathbf{u}_t}$ with feature maps, similarly to convolutional sparse coding \cite{yang2010supervised}. Chalasani and Principe \cite{chalasani2015context} used two layer  convolutional DPCNs
to extract features $\mathbf{u}_t$ that were then used to train SVMs for classification surpassing several other sparse auto-encoding
\cite{kavukcuoglu2010learning} and deconvolutional \cite{zeiler2010deconvolutional} techniques in accuracy on Caltech 101, Honda/UCSD faces \cite{KCLee03},
Celebrity Faces, and Coil-100 datasets.

Unfortunately, a few drawbacks exist that prevent DPCNs from scaling up and limit their application to other data
domains. The DPCN uses expectation maximization to search for the codes ${\mathbf{s}_t, \mathbf{u}_t}$ which best maximize the likelihood of input $\mathbf{x}_t$. This search is one of the reasons
for DPCN's good results, but it is also a drawback. Even during test time it is necessary to execute computationally
expensive expectation steps in order to estimate multidimensional coefficients ${\mathbf{s}_t, \mathbf{u}_t}$ for each layer. Here instead we propose
a parametrization using deep convolutional auto-encoders to compute the coefficients in a single operation for each layer.

Although successful as a model of the early stages of the visual system, imposing the multiplicative constraint $\mathbf{z}_t (k)$ limits the types
of features learned in the higher layers of the architecture. In fact, previous experiments with DPCNs failed to learn better features with more than two layers, which is not usually the case
for other deep learning architectures either supervised \cite{krizhevsky2012imagenet} or unsupervised \cite{lee2009convolutional}.

Here instead, we will focus on learning deep representations
from data using stacked rectified linear unity (ReLU) layers. This allows for simpler model building blocks which are still
powerful enough for vision \cite{krizhevsky2012imagenet} and can be possibly extended to other datasets, such as audio \cite{zeiler2013rectified}.

DPCNs assume that given an input $\mathbf{x}_t$, the current latent code $\mathbf{s}_t$ can be estimated given only the previous state $\mathbf{s}_{t-1}$ and $\mathbf{u}_t$.
Meanwhile, no backpropagation through time is carried out to update the parameters ${\mathbf{A}, \mathbf{B}, \mathbf{C}}$. Although this might be interesting for learning in
real-time without storing previous states, it limits the model from learning long-term dependencies. Also, without end-to-end connection
the model cannot be fully fine-tuned with supervised learning, which has been shown to improve results on auto-encoders \cite{hinton2006reducing}.

In the next section, the model we propose explicitly decouples spatial and temporal representations by learning state transitions in
the latent space with RNNs. We keep the smoothness in time by forcing the RNN transitions to stay in a fixed radius open ball around
the previous states.

Given DPCN's strengths and weaknesses, we attempt to develop an alternative architecture that is scalable, flexible and differentiable
through time, while keeping as much as possible DPCN's abilities to learn discriminative statistics from spatio-temporal context.

\section{Recurrent Winner-Take-All Network}
\label{section:rwta}

\begin{figure*}[t]
\centering
\includegraphics[scale=0.41]{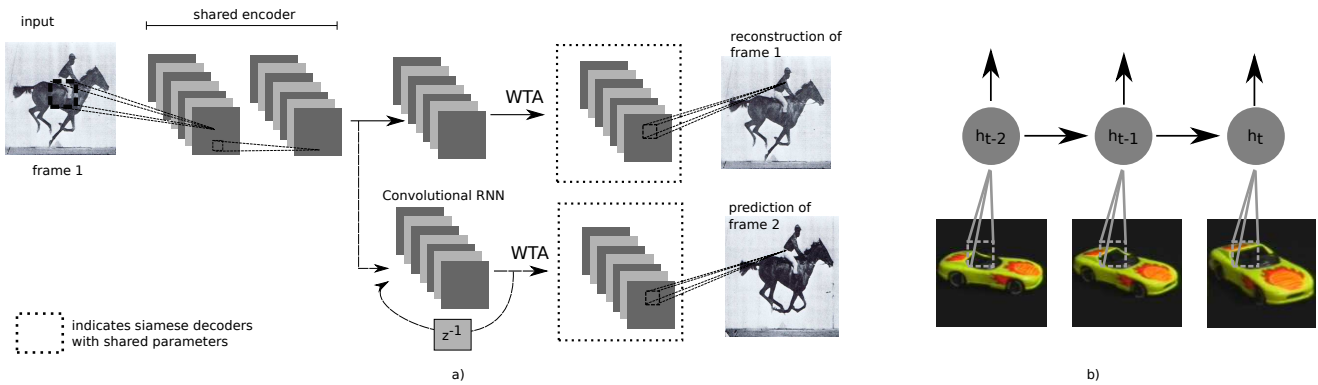}
\caption{Left: Schematic diagram of the proposed two-stream convolutional recurrent neural networks architecture with Winner-Take-All (RWTA) regularization. Upper stream is the static encoder-decoder. Lower stream denotes the temporal, dynamic encoder. Right: Representation of a convolutional recurrent neural network unfolded in time for a fixed spatial location. The same recurrent neural network is applied throughout the entire image.}
\label{fig:model}
\end{figure*}

In this section we propose an end-to-end differentiable convolutional-recurrent neural network with Winner-Take-All dropout for
feature extraction from video. In place of the linear state prediction matrix $A$, we use Convolutional Recurrent Neural Networks (ConvRNNs) to predict future states. RNNs are particularly appropriate for this framework as they have a long history \cite{graves2013generating} of successfully modeling dynamical systems. Furthermore, in place of using computationally expensive EM algorithms to compute the sparse states and causes, we use convolutional autoencoders with Winner-Take-All \cite{makhzani2015winner} regularization to encode the states in feedforward manner.

Convolutional-recurrent neural networks \cite{xingjian2015convolutional}\cite{liang2015recurrent} are RNNs where all the input-to-state and state-to-state transformations are implemented with convolutions.
A vanilla-RNN state update can be represented in an equation as
\begin{equation}
    \label{eq:vanilla}
    \mathbf{h}_t = f(\mathbf{ \mathbf{W} } \mathbf{h}_{t-1} + \mathbf{V} \mathbf{x}_t),
\end{equation}
where the dynamic state vector $\mathbf{h}_t$ represents the history of the time series up to time $t$, and $f$ is a nonlinearity such as the hyperbolic tangent or ReLU.
In a convolutional RNN, the dynamic state is a 3-way tensor
$\mathbf{h}_{t; f,r,c}$ with $f$ channels, $r$ rows and $c$ columns. The hidden-to-hidden transition operation is a convolutional kernel $\mathbf{W}_{o, f, r_w, c_w}$ with
$r_w < r$ and $c_w < c$, and $o = f$ output channels. Similarly, we have a convolutional kernel for the input-to-hidden transition $\mathbf{V}_{f, f_x, r_w, c_w}$, where
$f_x$ is the number of channels in the input $\mathbf{x}_{t, f_x, r, c}$.
\begin{equation}
\label{eq:vanilla_conv_rnn}
\mathbf{h}_{t; f, r, c} = f(\mathbf{W}_{f, f, r_w, c_w}\star \mathbf{h}_{t-1; f, r, c} + \mathbf{V}_{f, f_x, r_w, c_w}\star \mathbf{x}_{t; f_x, c, r}),
\end{equation}
where $\star$ denotes the multi-channel convolution operator used in deep learning:
\begin{equation}
\label{eq:conv}
(W\star h)_{fij} = \sum_{a,b,c}\mathbf{W}_{f,a, b, c} \mathbf{h}_{a, i-b, j-c}.
\end{equation}
A schematic representation of convolutional RNN is shown in Figure \ref{fig:model}.

RNNs can be trained in several ways for sequence prediction \cite{graves2013generating}\cite{sutskever2013training}; here we focus on training our ConvRNN to predict the next frame of the sequence.
Overfitting in conventional RNNs is avoided using bottleneck layers, where the RNN hidden state dimensions are smaller than the input. On the other hand,
as discussed in the DPCN section, for image analysis we want to expand the latent space dimensionality and avoid overfitting with
sparseness constraints. The advantages of high dimensional representations are formalized by Cover's theorem.

In autoencoders, sparsity can be imposed as a constraint on the objective function \cite{olshausen1996emergence}. Unfortunately, sparseness as measured by $L_1$ or $L_0$ norms is hard to
optimize, requiring elaborate techniques such as proximal gradients (ex. FISTA \cite{beck2009fast}) or learned approaches (ex. LISTA \cite{gregor2010learning}). Since we want
sparseness in the network outputs and not network weights, those techniques would also require optimization during test time, as done in the original formulation of DPCNs.
Recent research has shown that simple regularization techniques such as Dropout \cite{hinton2012improving} combined with ReLU activations are enough to learn sparse representations
without extra penalties in the cost functions. This represents a paradigm shift from cost function to architectural regularization that provides faster training
and testing. 

Makhzani and Frey proposed Winner-Take-All (WTA) Autoencoders \cite{makhzani2015winner} which use aggressive Dropout, where all the elements but the strongest of a convolutional map are zeroed out. This forces
sparseness in the latent codes and the convolutional decoder to learn robust features. Here we extend convolutional Winner-Take-All autoencoders through time using convolutional RNNs. WTA for a map $x_{f, r, c}$ in the output of convolutional layer
can be expressed as in (\ref{eq:wta}). The indices $f, r, c$ represent respectively the number of rows, the number of columns, and the number of channels in the map.

\begin{equation}
\label{eq:wta}
WTA(x_{f, r, c}) = \begin{cases} x_{f, r, c}, \quad \text{if} \quad x_{f, r, c} = \underset{r, c}{\max}(x_{f, r, c}) \\ 0, \quad \text{otherwise} \end{cases}.
\end{equation}

Thus, $WTA(x_{f, r, c})$ has only one non-zero value for each channel $f$. To backpropagate through (\ref{eq:wta}) we use $\nabla WTA(x_{f, r, c}) = WTA(\nabla x_{f, r, c})$. In the present paper, we apply (\ref{eq:wta}) to the output of the convolutional maps
of the ConvRNNs after they have been calculated. In other words, the full convolutional map hidden state is used inside the dynamics of the ConvRNN, WTA is applied only before they are fed as input to the convolutional decoder. We leave investigations of 
how WTA would affect the inner dynamics of ConvRNNs for future work.

We propose to learn smoothness in time with architectural constraints using a two-stream encoder as shown in Figure \ref{fig:model}.
This architecture was inspired by the dorsal and ventral streams hypothesis in the human visual cortex \cite{goodale1992separate}. Roughly speaking, the dorsal stream models "vision for action"
and movements and the ventral stream represents "vision for perception". In our proposed architecture one stream is a stateless convolutional encoder-decoder and
the other stream has a convolutional RNN encoder, thus a dynamic state. Using Siamese decoders for both streams, we force the stateless encoder and the convolutional RNN to
project into the same space---one which can be reconstructed by the shared weights decoder. It is important to stress that from the point of view of spatio-temporal feature extraction with the ConvRNN, the stateless stream works as for regularization. As any other sort of regularization its usefulness can only be totally stated in practice and the practitioner might optionally not use it. Nevertheless, we opted for using the full architecture in all the experiments of this paper. In Appendix A, we show how this proposed architecture enforces spatio-temporal smoothness in the embedded space.

\begin{figure*}[t]
\centering
\includegraphics[scale=0.52]{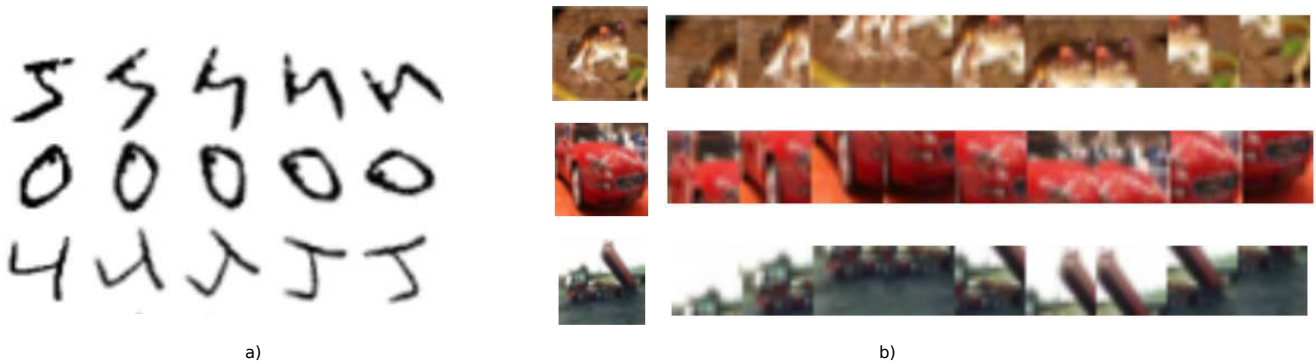}
\caption{Left: Sample videos of the rotated MNIST dataset. Right: Sample original images and videos generated by scanning the images with 16x16 pixels windows. The scanning runs vertical first, starting in the upper left corner.}
\label{fig:mnist}
\end{figure*}

Given an input video stream $\mathbf{x}_t$, denoting the stateless encoder by $E$, the decoder $D$, and the convolutional RNN by $R$,
the cost function for training our architecture is the sum of reconstruction and prediction errors:
\begin{equation}
    \label{eq:cost}
    L_t = \mathbb{E}\left[(\mathbf{x}_{t-1} - D(E(\mathbf{x}_{t-1})))^2 + (\mathbf{x}_{t} - D(R(\mathbf{x}_{t-1})))^2 \right],
\end{equation}
where $\mathbb{E}$ denotes the expectation operator. Notice that as depicted in Figure \ref{fig:model}, $E$ and $R$ have shared parameters.
During training, we observe a few input frames $t = [1, 2, ..., T]$ and adapt all the parameters
using backpropagation through time (BPTT) \cite{werbos1990backpropagation}. Notice that due to BPTT both streams of our architecture are adapted while considering temporal context. Thus,
the stateless encoder $E$ will learn richer features than it would if trained on individual frames.

As great power brings great responsibility, the main drawback of our proposed architecture is the memory required by BPTT and convolutions. The
gradients of convolutions require storing the multi-way tensor output of the convolutions, and BPTT requires storing all the outputs for all time steps.
The combination of both methods in a single architecture requires powerful hardware. We limited the length of our input time series between 5 and 10 frames, which is also the length used by the methods with which we compare in the Experiments section. In the next section, we compare our proposed architecture with similar methods proposed in the literature beyond the already discussed DPCN.

\section{Related Work}
This research is related to DPCNs and a larger family of deep unsupervised neural networks \cite{olshausen1996emergence}\cite{noroozi2016unsupervised}\cite{lee2009convolutional}.
The aforementioned Winner-Take-All Autoencoders (WTA-AE) \cite{makhzani2015winner} consist of a deep convolutional encoder and a single layer convolutional decoder,
which inspired our choice. WTA-AE drops out all the elements of a convolutional channel map but the largest, forcing the whole system to learn robust, sparse features for reconstruction.
With the proposed convolutional RNN our method can be seen as a natural extension of WTA-AE.

Unsupervised learning with temporal context was also previously explored by Goroshin et. al. \cite{Goroshin_2015_ICCV} Wang and Gupta \cite{wang2015track}. Their approach
was based on metric learning of related frames in video, but their approaches were not capable of learning long term dependencies since they assumed only a simple
zero-mean Gaussian innovation between frames. Also, neither of these approaches can be fine-tuned by BPTT
to learn end-to-end classifiers in time.

Convolutional RNNs were proposed simultaneously by several authors \cite{liang2015recurrent}\cite{xingjian2015convolutional} as an extension of
Network-in-Networks \cite{nin} architectures where each convolutional layer of a CNN are themselves deep networks. Liang et. al \cite{liang2015recurrent}
proposed to make each convolutional layer a fixed input RNN. They used that architecture for object recognition in static
scenes without exploring temporal context. Xingjian et. al. \cite{xingjian2015convolutional} and Patraucean et. al. \cite{patraucean2015spatio}
on the other hand used temporal context in videos for weather and video forecasts. Their architectures
are similar to predictive networks, but they did not address the problem of regularizing the latent space features, nor how to train deep architectures---
their models consist only of a single convolutional RNN module for predicting future frames. Furthermore, they do not investigate how to extract interesting features
without context, which is the problem addressed by our stateless encoder-decoder stream trained in parallel with the dynamic stream.

In parallel to this paper, another follow up on the DPCN approach was published by Bill Lotte et. al \cite{prednet}. Differently from this paper, 
their method, called PredNet, focused on frame prediction for video and not sparse feature extraction. Nevertheless, both approaches are 
complementary and could be combined in future work. 

\section{Experiments}
To illustrate the capabilities of our proposed architecture we applied it first to two artificial datasets generated by modifying the MNIST and Cifar10 datasets.
We used MNIST and Cifar10 as development datasets to understand how hyperparameter choices affect our method; that is, to understand how many filters per layer are
necessary, how much temporal context contributes to learning unsupervised features, and how long to take in the unsupervised phase. The full list of hyperparameters is shown on Table \ref{table:hp}. 
Note that we fixed the number of channels per convolutional layer be equal in layers to limit the number of hyperparameters. An exception is the number of channels in the decoder (the very last layer) since it has to match the number of channels 
in the input (i.e. 1 for black and white images and 3 for color images).

Furthermore, for the modified MNIST dataset, we show our architecture learns, using temporal information, more discriminative features. For the modified Cifar10 dataset, we show the advantage of pre-training convolutional RNNs with our method.

Afterwards, we applied our best performing architectures to the Coil100 and Honda/UCSD Faces Dataset for a direct comparison with DPCN and other unsupervised
learning techniques.

\begin{table*}
\caption{Hyperparameter choices per experiment}
\center
\begin{tabular}{ccccc}
\hline
 & rotated MNIST & scanned Cifar10 & COIL100 & Honda Faces \tabularnewline
\hline
Channels per layer & 64 & 256 & 128 & 256 \tabularnewline
Filter size (encoder) & 3x3 & 5x5 & 5x5 & 5x5\tabularnewline
Filter size (decoder) & 11x11 & 7x7 & 7x7 & 7x7 \tabularnewline
\hline
\multicolumn{5}{c}{All models were trained using ADAM optimization rule with learning rate 0.001} \tabularnewline
\multicolumn{5}{c}{All models were 4 layers deep} \tabularnewline
\multicolumn{5}{c}{All models had 2 convolutional layers before the ConvRNN layer.} \tabularnewline
\multicolumn{5}{c}{WTA was applied only right before the last layer.} \tabularnewline
\hline
\end{tabular}
\label{table:hp}
\end{table*}

\subsection{Rotated MNIST Dataset}
We extended the MNIST dataset by generating videos by rotating each image counter-clockwise. Sample videos are shown in Fig. \ref{fig:mnist}. We trained our two-stream convolutional RNN on videos generated with MNIST training dataset. The task was to learn
to reconstruct and predict frames as described in \ref{eq:cost}. We trained the networks with batches of size 100 for 3 epochs (a total of 1800 updates) using the Adam learning rule \cite{kingma2014adam}.
We trained a linear Support Vector Machine (SVM) on the features computed by the convolutional RNN $R$. We collapsed the temporal features into one
using addition: $z = \sum_t R(x_t)$, where the $z$'s are the input to train the SVM. All the encoder convolutional kernels had $f=64$ channels of size $c=r=3$.
The classification error probability on videos generated with the MNIST test set was $0.94\%$. An equivalent WTA-AE obtained only $1.02\%$ accuracy.

We argue that the possible reasons for the better performance of the proposed method are due to data augmentation and the capacity of the proposed method
to use that augmentation to compose a single, less ambiguous interpretation of the data. In Figure \ref{fig:mnist} we show the 64 filters of 11x11 pixels
learned by the decoder $D$. 

\subsection{Scanned Cifar10 dataset}
The Cifar10 dataset consists of 50k RGB images of 32x32 pixels for training and an additional 10k images for testing. There is a total of 10 different classes. We converted this dataset
to videos by scanning it with 16x16 windows that move 8 pixels at a time as shown in Fig. \ref{fig:mnist}. The 16x16 windows were pre-processed with ZCA.
In most of the videos no single 16x16 window completely captures
the object to be classified. This forces a classifier to use "temporal" context to perform well.

We trained the proposed method on this dataset for 10 epochs. Each convolutional map of the encoders $E$ and $R$ had 256 filters of size 5x5 pixels. The decoder had filters of
size 7x7. We then fine-tuned the convolutional RNN for classification using supervised learning and obtained a classification rate of $75.6\%$, while a similar convolutional RNN trained from
scratch obtained only $74.1\%$. Both networks were equally initialized using the Glorot uniform method \cite{glorot2010understanding}.

We did not have success using a single linear SVM on sum-collapsed features
for this dataset. Nevertheless, this experiment suggests that even when the proposed pre-training technique in itself is not enough for learning relevant features, it can still be
used as an initialization technique for complex recurrent systems. 

Using what we learned with these two preliminary examples on the modified MNIST and Cifar10 datasets we decided to use the following general guidelines for the following experiments: 1) Color videos are preprocessed by ZCA. 2) Convolutional filters use either 128 or 256 channels of size 3x3 or 5x5 in the encoder and 7x7 in the decoder. 3) Training takes $\approx 1500$ updates. For our dataset this was about 10 epochs long, with batch sizes of 16 or 32, depending on GPU memory. 4) Linear SVM classifiers were trained on encoded version of the last frame of a sequence $R(x_T)$. In our experiments $T=5$. For longer videos in test time, we classified each frame using by moving the $T=5$ window one frame at a time and took the most voted class as the final guess.

\subsection{Coil-100 Dataset}
The COIL-100 dataset \cite{nene1996columbia} consists of 100 videos of different objects. Each video is 72 frames long and were generated by placing the object on a turn table and taking
a picture every $5^{\circ}$. The pictures are 128x128 pixels RGB. For our experiments, we rescaled the images to 32x32 pixels and used ZCA pre-processing.

\begin{figure}[t]
\centering
\includegraphics[scale=0.5]{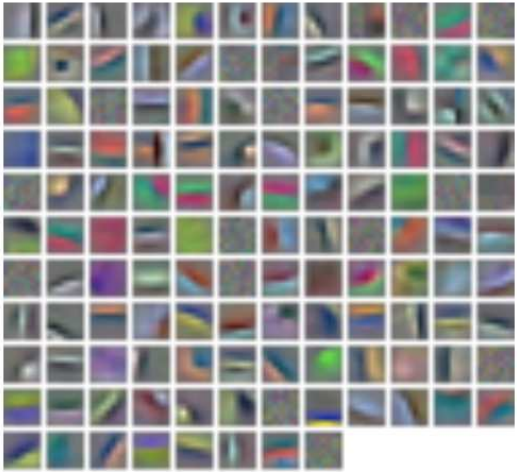}
\caption{128 decoder weights of 7x7 pixels learned on Coil-100 videos.}
\label{fig:coil}
\end{figure}

The classification protocol proposed in the COIL-100 \cite{nene1996columbia} uses 4 frames per video as labeled samples, the frames corresponding to angles $0^{\circ}, 90^{\circ}, 180^{\circ}$ and $270^{\circ}$.
Chalasani and Principe\cite{chalasani2015context} and Mobahi et. al. \cite{mobahi2009deep} used the entire dataset for unsupervised pre-training. For this reason, we believe the results in this experiment should be understood with this in mind. Note that the compared methods enforce smoothness in the representation of adjacent frames, and since the test frames are observed in context for feature extraction, information
is carried from labeled to unlabeled samples. In other words, this experiment is better described as semi-supervised metric learning than unsupervised learning. Here, we followed that same
protocol, using 14 frames per video. Results are reported in Table \ref{table:coil}. We used encoders with 128 filters of 5x5 pixels and a decoder with 7x7 pixels. The decoder filters are shown in Fig. \ref{fig:coil}

\begin{table}
\caption{Recognition rate (in percentage \%) for object recognition in Coil-100 dataset}
\center
\begin{tabular}{cc}
\hline
Method & Accuracy\tabularnewline
\hline
DPCN no context \cite{chalasani2015context} & 79.45\tabularnewline
Stacked ISA + temporal \cite{le2011learning} & 87\tabularnewline
ConvNets + Temporal \cite{mobahi2009deep} & 92.25 \tabularnewline
DPCN + temporal + top down \cite{chalasani2015context} & 98.34\tabularnewline
Proposed method & 99.4\tabularnewline
\hline
\end{tabular}
\label{table:coil}
\end{table}

\begin{table*}
\caption{Recognition rate (in percentage \%) for face recognition in Honda/UCSD dataset}
\center
\begin{tabular}{ccccc}
\hline
Sequences Lengths & MDA \cite{wang2009manifold} & SANP \cite{hu2011sparse} & CDN \cite{chalasani2015context} & \begin{tabular}{@{}c@{}}Proposed \\ Method\end{tabular} \tabularnewline
\hline
50 Frames & 74.36 & 84.62 & 92.31 & 100\tabularnewline
100 Frames & 94.87 & 92.31 & 100 & 100\tabularnewline
Full Video & 97.44 & 100 & 100 & 100\tabularnewline
\hline
\end{tabular}
\label{table:honda}
\end{table*}

\subsection{Honda/UCSD Dataset}
The Honda/UCSD dataset consists of 59 videos of 20 different people moving their heads in various ways. The training set consists of 20 videos (one for each person), $\sim 300-1000$ frames each. The test set consists of 39 videos (1-4 per person), $\sim 300-500$ frames each. For each frame of all videos, we detected and cropped the faces using Viola-Jones face detection. Each face was then converted to grayscale, resized to 20x20 pixels, and histogram equalized.

During training, the entire training set was fed into the network, 9 frames at a time, with a batch size of 32. After training was complete, the training set was again fed into the network. For each input frame in the sequence, the feature maps from the convolutional RNN were extracted, and then (5,5) max-pooled with a stride of (3,3). In accordance with the test procedure of Chalasani and Principe \cite{chalasani2015context}, a linear SVM was trained using these features and labels indicating the identity of the face. Finally, each video of the test set was fed into the network, one frame at a time, and features were extracted from the RNN in the same way as described above. Each frame was then classified using the linear SVM. Each sequence was assigned a class based on the maximally polled predicted label across each frame in the sequence. Table \ref{table:honda} summarizes the results for 50 frames, 100 frames, and the full video, comparing with 3 other methods, including the original convolutional implementation of DPCN \cite{chalasani2015context}. The results for the 3 other methods were taken from \cite{chalasani2015context}. The results for our method were perfect for all the tested cases.

\section{Preliminary results and future work}
In all our experiments we investigate single scale feature extraction, i.e., we did not use pooling or strided convolutions. In experiments not discussed in this paper, we explored
pooling and unpooling in the proposed architecture. However, this did not improve the results considerably. Nevertheless, Makhzani and
Frey \cite{makhzani2015winner} showed that layer wise training at different scales improved their results. They first trained a convolutional WTA-AE on the raw data, downsampled the
features with maxpooling and trained another layer on top the pooled features. Learning at a different scale could be done in two different ways with the proposed architecture. The first way would be by collapsing
the temporal features into a single feature map (by addition or picking the last state), downsampling and learning a WTA-AE on top for the resulting features. The second approach would be
to downsample every feature in time and train a second two-stream convolutional RNN. Which approach is best remains elusive, but we plan to investigate further in future work.

In our experiment sections, we showed how the proposed architecture compares favorably to other methods that leverage temporal context for object recognition. Future experiments should
focus on problems such as action recognition, where the best performance cannot be achieved using single frame recognition. In preliminary experiments, we applied the same architecture
used with the Coil-100 dataset to the UCF-101 action recognition dataset. A linear SVM applied to unsupervised features was used to obtain a recognition rate of $44\%$ which is slightly
better than the baseline of $43.90\%$. On the other hand, methods based on supervised learning with deep convolutional neural networks using hand-engineered spatial flow features can
obtain a recognition rates $>80\%$ \cite{simonyan2014two}\cite{tran2014learning}. Our follow up work should investigate how to learn unsupervised features with architectural constraints similar to those used for
calculating spatial flow, e.g. local differences in the pixel space.

When investigating the benefits of unsupervised training for discriminative convolutional RNNs, we only showed results for unsupervised initialization.
However, Luong et. al. \cite{Luong2015multitask} showed that multitask learning can improve the overall results of the supervised learning task. In other words, while training a conventional RNN
for classification, they also added an unsupervised term to the objective function. They argued that such multitask learning regularized the RNN and avoided overfitting. In convolutional
architectures the number of latent features (outputs of convolutional layers) is much larger than the number of learned parameters, which already enforces some regularization. However,
given that natural images are highly correlated in local neighborhoods, we believe that multitask learning will also benefit convolutional RNNs trained for supervised tasks.

\section{Conclusions}
\label{sec:conclusions}
This paper proposes RWTA, a deep convolutional recurrent neural network with Winner-Take-All dropoutfor extracting features from time series. Our contributions were threefold: 1) a scalable, end-to-end differentiable
reinterpretation of the sparse spatio-temporal feature extraction in Deep Predictive Coding Networks \cite{chalasani2015context}; 2) an extension of Winner-Take-All Autoencoders \cite{makhzani2015winner} to time using dynamic neural networks; 3) a new technique for
initializing and regularizing \cite{Luong2015multitask} convolutional recurrent neural networks \cite{Liang_2015_CVPR}\cite{shi2015convrnn}. We showed that our method outperforms DPCNs
and other similar methods in contextual object recognition tasks. We also showed that this method can be used as an initialization technique for supervised convolutional RNNs, obtaning better
results than Glorot initialization \cite{glorot2010understanding}.

\appendices
\section{RWTA learned invariances}

\newcommand{\R}{\mathbb{R}}
\newcommand{\N}{\mathbb{N}}
\newcommand{\Z}{\mathbb{Z}}
\newcommand{\Q}{\mathbb{Q}}
\newcommand{\C}{\mathbb{C}}
\newcommand{\K}{\mathbbm{K}}

\newcommand{\mRe}{\ensuremath{\mathbb{R}}}

\providecommand{\norm}[1]{\left\|#1\right\|}
\newcommand{\conv}{\xrightarrow[n\to\infty]}

\newcommand{\mmb}[1]{\mathbf{#1}}
\newcommand{\mmbs}[1]{\mathbf{\scriptstyle{#1}}}
\newcommand{\mmbss}[1]{\mathbf{\scriptscriptstyle{#1}}}

\newcommand{\mPr}[1]{\mathcal{P}\left\{#1\right\}}

\newcommand{\mdf}[2][f]{#1_{#2}}
\newcommand{\mdfp}[3][f]{#1_{#2;#3}}
\newcommand{\mdfa}[3][f]{#1_{#2}(#3)}
\newcommand{\mdfpa}[4][f]{#1_{#2;#3}(#4)}

\newcommand{\mEx}[2]{\mathbb{E}_{\scriptscriptstyle{#1}}\left\{#2\right\}}

\newcommand{\shannon}[1]{\mathrm{h}_{\scriptscriptstyle{\mathrm{S}}}\left(#1\right)}
\newcommand{\crossentropy}[2]{\mathrm{h}_{\scriptscriptstyle{\mathrm{CE}}}\left(#1,#2\right)}
\newcommand{\kullbackleiblerd}[2]{\mathrm{d}\,_{\scriptscriptstyle{\mathrm{KL}}}\left(#1;#2\right)}

\newcommand{\renyi}[2]{\mathrm{h}_{\scriptscriptstyle{#1}}\left(#2\right)}
\newcommand{\renyid}[3]{\mathrm{d}_{\scriptscriptstyle{#1}}\left(#2;#3\right)}

\newcommand{\fscore}[2]{\mathrm{v}_{\scriptscriptstyle{\mathrm{F}}}\left(#1\right)_{\scriptscriptstyle{#2}}}
\newcommand{\fisher}[2]{\mathrm{j}_{\scriptscriptstyle{\mathrm{F}}}\left(#1\right)_{\scriptscriptstyle{#2}}}
\newcommand{\crossinformation}[3]{\mathrm{j}_{\scriptscriptstyle{\mathrm{CI}}}\left(#1,#2\right)_{\scriptscriptstyle{#3}}}
\newcommand{\informationcorrelation}[2]{\mathrm{j}_{\scriptscriptstyle{\mathrm{ICO}}}\left(#1\right)_{\scriptscriptstyle{#2}}}
\newcommand{\rfitI}[3]{\mathrm{d}\,_{\scriptscriptstyle{\mathrm{J(I)}}}\left(#1;#2\right)_{\scriptscriptstyle{#3}}}
\newcommand{\rfitII}[3]{\mathrm{d}\,_{\scriptscriptstyle{\mathrm{J(II)}}}\left(#1;#2\right)_{\scriptscriptstyle{#3}}}

\newcommand{\scorei}[3]{\mathrm{v}_{\scriptscriptstyle{\mathrm{(i)}}}\left(#1\right)_{\scriptscriptstyle{#2};\scriptscriptstyle{#3}}}
\newcommand{\informationi}[3]{\mathrm{j}_{\scriptscriptstyle{\mathrm{(i)}}}\left(#1\right)_{\scriptscriptstyle{#2};\scriptscriptstyle{#3}}}

\newcommand{\scoreii}[3]{\mathrm{v}_{\scriptscriptstyle{\mathrm{(ii)}}}\left(#1\right)_{\scriptscriptstyle{#2};\scriptscriptstyle{#3}}}
\newcommand{\informationii}[3]{\mathrm{j}_{\scriptscriptstyle{\mathrm{(ii)}}}\left(#1\right)_{\scriptscriptstyle{#2};\scriptscriptstyle{#3}}}

\newtheorem{lemma}{Lemma}

Assume input stream $\mmb{x}_t$, where $t$ is a countable index, is fed into two modules, a static and a recurrent neural network respectively:
\begin{eqnarray}
\mmb{o}_t^E & = & \mmb{e}(\mmb{x}_t)\\
\mmb{o}_t^R & = & \mmb{r}(\mmb{x}_t, \mmb{o}_{t-1}),
\end{eqnarray}
where $\mmb{r}$ is a recurrent neural network and the state  $\mmb{o}_{t-1}$ gives it context for prediction. These two outputs are fed into a siamese decoders that produces another two outputs
\begin{eqnarray}
\mmb{f}_t^E & = & \mmb{d}(\mmb{o}_t^E)\\
\mmb{f}_t^R & = & \mmb{d}(\mmb{o}_t^R)
\end{eqnarray}
Training is done such that the following expression is minimized:
\begin{equation}
E = \sum_{t}{(\mmb{x}_{t-1}-\mmb{f}_{t-1}^E)^2}+\sum_{t}{(\mmb{x}_t-\mmb{f}_{t-1}^R)^2}
\end{equation}

The main effect of the WTA algorithm \cite{Makhzani2015} when applied to $\mmb{o}_t^E$ and $\mmb{o}_t^R$, is to partition the input space of the corresponding functions into volumes that produce the same output. Hence, if there is a sequence of inputs in the interval $\{t,\ldots,t+k\}$ that is contained inside one of these volumes, then:

\begin{equation}
\begin{split}
\mmb{o}^E = \mmb{o}_t^E = \mmb{o}_{t+1}^E = \cdots = \mmb{o}_{t+k}^E = \mmb{e}(\mmb{x}_t) = \\
\mmb{e}(\mmb{x}_{t+1}) = \cdots = \mmb{e}(\mmb{x}_{t+k})
\end{split}
\end{equation}

\begin{equation}
\begin{split}
\mmb{o}^R = \mmb{o}_t^R = \mmb{o}_{t+1}^R = \cdots = \mmb{o}_{t+k}^R =\mmb{r}(\mmb{x}_t) = \\
 \mmb{r}(\mmb{x}_{t+1}) = \cdots = \mmb{r}(\mmb{x}_{t+k})
\end{split}
\end{equation}

which implies
\begin{eqnarray}
\mmb{f}^E & = & \mmb{f}_t^E = \mmb{f}_{t+1}^E = \ldots = \mmb{f}_{t+k}^E = \mmb{d}(\mmb{o}^E) \\
\mmb{f}^R & = & \mmb{f}_t^R = \mmb{f}_{t+1}^R = \ldots = \mmb{f}_{t+k}^R = \mmb{d}(\mmb{o}^R)
\end{eqnarray}

Hence, isolating the corresponding section of the objective function, and using the previous equalities, the following expression needs to be minimized:
\begin{eqnarray}
E_{t,t+k} & = & (\mmb{x}_t-\mmb{f}_t^E)^2+(\mmb{x}_{t+1}-\mmb{f}_t^R)^2 \nonumber \\
& & +(\mmb{x}_{t+1}-\mmb{f}_{t+1}^E)^2+(\mmb{x}_{t+2}-\mmb{f}_{t+1}^R)^2 \nonumber \\ & & \cdots \nonumber \\
& & +(\mmb{x}_{t+k}-\mmb{f}_{t+k}^E)^2+(\mmb{x}_{t+k+1}-\mmb{f}_{t+k}^R)^2 \\
& = & (\mmb{x}_t-\mmb{f}^E)^2+(\mmb{x}_{t+1}-\mmb{f}^R)^2 \nonumber \\
& & +(\mmb{x}_{t+1}-\mmb{f}^E)^2+(\mmb{x}_{t+2}-\mmb{f}^R)^2 \nonumber \\
& & \cdots \nonumber \\
& & +(\mmb{x}_{t+k}-\mmb{f}^E)^2+(\mmb{x}_{t+k+1}-\mmb{f}^R)^2 \\
& = & (\mmb{x}_t-\mmb{d}(\mmb{o}^E))^2+(\mmb{x}_{t+1}-\mmb{d}(\mmb{o}^R))^2 \nonumber \\
& & +(\mmb{x}_{t+1}-\mmb{d}(\mmb{o}^E))^2+(\mmb{x}_{t+2}-\mmb{d}(\mmb{o}^R))^2 \nonumber \\
& & \cdots \nonumber \\
& & +(\mmb{x}_{t+k}-\mmb{d}(\mmb{o}^E))^2+(\mmb{x}_{t+k+1}-\mmb{d}(\mmb{o}^R))^2
\end{eqnarray}

Thus we can do the following considerations. Minimization of the last expression shows that $\mmb{d}(\mmb{o}^E)$ and  $\mmb{d}(\mmb{o}^R)$ will try to be close to all the $\mmb{x}_t, \mmb{x}_{t+1}, \ldots, \mmb{x}_{t+k}$ in a square sense. Hence it will produce $\mmb{f}_{\star}^E = \mmb{d}(\mmb{o}^E)$ and  $\mmb{f}_{\star}^R = \mmb{d}(\mmb{o}^R)$.
Moreover, neglecting the two extreme terms of the last expression shows that it is composed of pairs of sums such as $(\mmb{x}_{t+1}-\mmb{d}(\mmb{o}^R))^2+(\mmb{x}_{t+1}-\mmb{d}(\mmb{o}^E))^2$. Thus the minimization process is also trying to make $\mmb{d}(\mmb{o}^R) \approx \mmb{d}(\mmb{o}^E)$ which implies $\mmb{o}^R \approx \mmb{o}^E$. This forces a balance between the stateless encoder and the recurrent neural network. Hence the system will look for solution that consider spatial and temporal invariances.
Given that the capacity of the system to partition the space is limited, there is a limited number of $\mmb{o}^R$ and $\mmb{o}^E$ encodings that the system can produce. Thus the training procedure will focus on finding those volume partitions that can be used to explain the incoming stream in an efficient manner.
The combined effect of all these components finally act like an invariance detector.


\end{document}